# Graph based Label Enhancement for Multi-instance Multi-label learning

**Houcheng Su[1], Jintao Huang[2], Daixian Liu[3], Rui Yan[3], Jiao Li[3], Chi-Man Vong[2,*]**

[1]Institute of Collaborative Innovation, University of Macau, Macau S.A.R
[2]Faculty of Science and Technology, University of Macau, Macau S.A.R
[3]College of Information Engineering, Sichuan Agricultural University, China
{mc25695, hjt.violler }@connect.um.edu.mo, {202105787,
202105891,202005852}@stu.sicau.edu.cn,cmvong@um.edu.mo

## Abstract

Multi-instance multi-label (MIML) learning ignores label significance, such as the image classification where one image contains multiple semantics as instances, correlated with multiple labels of different significance in real are reduced to the logical labels. Ignoring labeling significance will greatly lose the semantic information of the object, so that MIML is not applicable in complex scenes with a poor learning performance. However, existing LE methods focuses on single instance tasks in which the inter-instance information in MIML is ignored, numerous feature spaces of MIML makes traditional LE difficult to mine implicit information simultaneously. To this end, this paper proposed a novel MIML framework based on graph label enhancement, namely GLEMIML, to improve the classification performance of MIML by leveraging label significance. GLEMIML first recognizes the correlations among instances by establishing the graph and then migrates the implicit information mined from the feature space to the label space via nonlinear mapping, thus recovering the label significance. Finally, GLEMIML is trained on the enhanced data through matching and interaction mechanisms. GLEMIML (AvgRank: 1.44) can effectively improve the performance of MIML by mining the label distribution mechanism and show better results than the SOTA method (AvgRank: 2.92) on multiple benchmark datasets.

## 1 Introduction

Objects in the real world are often polysemantic, consisting of multiple instances associated with multiple labels [Zhou, 2012]. Traditional supervised learning can be regarded as the degradation of objects with complex semantics, where useful information is lost at the representation stage. Nevertheless, in the multi-instance multi-label learning (MIMLL) framework, the objects can correspond to a bag of instances with a set of labels. In MIMLL, numerous practical problems can be properly formalized [Zhang and Wang, 2009]. For example, multi-structural domain multifunctional proteins exist in nature that are formed by aggregating multiple structural domains. Structural domains may perform their functions independently or perform multiple functions in concert with neighbouring structural domains [Wu et al., 2014]. In the MIMLL, different structural domains can be divided into different instances and form the multifunctionality of a protein into a set of functional labels that can more effectively represent real-world problems.

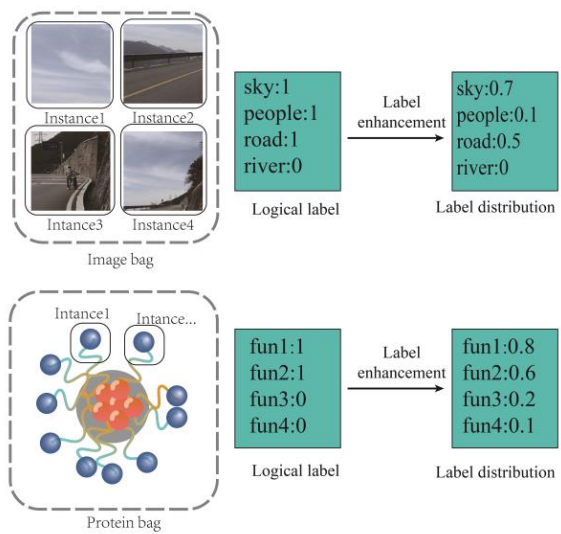

Figure 1: An example of label enhancement for MIML learning

In MIML tasks, simplified logical labels such as {0,1} are often used for labeling, thus losing more abundant semantic information. As shown in Figure 1, the significance of logical labels in each bag differs significantly. The above situations abound in practical applications. If ignoring the influence of labeling significance, MIML will be inaccurate and ineffective. To this end, it is urgent to leverage the labeling significance with richer semantic information from the existing logical labels of MIML.

Label enhancement (LE) is a tool to efficiently mine label significance (i.e., also known as label distribution[Geng, 2016]). Nevertheless, existing LE methods mainly focus on single instance multiple labels (SIML) tasks [Zhang et al., 2020]. If applying LE directly to MIML, a downgrade strategy may be required to convert MIML tasks to SIML tasks, thus using LE. However, this will lead to a significant loss of important implicit information based on inter-instance correlations. In addition, there may be numerous redundant features in MIML, resulting in traditional label enhancement with noisy information, thus cannot effectively utilize the recovered label distribution information. With difficulty in label quantification, it is challenging to construct the label distribution data artificially. Consequently, it is imminent to mine the potential label distribution information of MIML effectively.

To this end, this paper applies the features embedding for in-bag instances to mine the inter-instances correlation via Laplace matrices. To cope with the large and indeterminate feature space, the topological information of the feature space is mined by mapping it to a common subspace. Subsequently, the migrated information is used to recover the label distribution, and the inter-label correlation information is mined using Laplace matrices in the label distribution space. Finally, to avoid the label space recovered using the large feature space not being effectively exploited by the MIML classifier, a matching interaction mechanism is used to match LE with a multi-instance label distribution classifier through a hybrid label loss function. With the lack of a multi-instances label distribution dataset, we apply this to the MIML dataset and compare it with various approaches.

The main contributions of this paper are as follows: 1) a new MIML label enhancement method, GLEMIML, is proposed, to provide new insights into label enhancement in a multi-instance framework. 2) A new method for mining inter-instances correlation information is proposed to mine different instance by projecting them in a subspace utilizing successive embeddings.

## 2 Related Work

### 2.1 Multi-instance Multi-label learning

MIML learning has been extensively studied in recent years. Compared with traditional frameworks, MIML can solve problems more naturally in complex objects with multiple semantics [Zhou, 2012]. At the same time, MIMLBOOST and MIMLSVM algorithms based on degradation strategy and D-MIMLSVM algorithm based on regularization framework are proposed for the first time. After MIML framework was proposed, a MIML classification network was proposed by using MultiLayer Perceptrons (MLP) and optimized by using Back-Propagation (BP), which was called MIMLNN [Chen et al., 2013]. By improving on MIMLNN, combined with three Hausdorff distance metric optimizations, it is proposed that EnMIMLNN can show excellent performance in various protein data sets [Wu et al., 2014]. DeepMIML [Feng and Zhou, 2017] utilizes deep learning research to automatically locate key input forms that trigger labels while retaining MIML's instance-label relationship discovery capabilities. By combining CNN with MIML, a deep multimodal CNN for MIML image classification is proposed. By using the CNN architecture to automatically generate MIML instance representations, labeling is grouped by subsequent layers to achieve label correlation. Combined with label group context multiple Modal instances come from campuses to distinguish different groups of visually similar objects [Song et al., 2018].

### 2.2 Label Enhancement

Label enhancement aims to recover the label distribution from logical labels in the training set to guide classifiers' learning effectively. Graph Laplacian label enhancement (GLLE) [Xu et al., 2019] exploits the general topological information of the feature space and the correlation between labels to mine the hidden labeling significance. TMV-LE [Zhang et al., 2020] utilizes the factorization of tensors to adopt general representation with multi-view joint mining for a more comprehensive topology, which is to obtain the joint subspace of multi-view and migrate it to the label space. FLEM [Zhao et al., 2022] designed a matching and interaction mechanism, which completed the label distribution and prediction model correspondence with an integrated training process of LE. Label enhancement with Sample Correlations (LESC) uses the sample low-rank representation of the feature space. Generalized Label Enhancement with Sample Correlations (gLESC) uses tensor multi-rank minimization to explore the sample correlation in the feature space and label space [Zheng et al., 2021]. The Label Distribution based Confidence Estimation (LDCE) [Liu et al., 2021] estimates the confidence of the observed label, which cleans the boundary between the label and the noise label so that the reliable label can recover the label distribution.

## 3 Methodology

### 3.1 Notations Definition And Overall Framework

The related concepts and notations used in this paper are listed as follows. Given a MIML dataset consisting of the global feature space of all bags that are denoted as $X$. The feature space of $i$-th bag can be denoted as $x_i$, each bag has a different number of instances, $k$-th instance of $i$-th bag can be denoted as $x_i^k$. The feature space of $i$-th bag can be denoted as $x_i=\{x_i^1,...,x_i^k\}$, so the global feature space of all bags $X$ also can be expressed as $X=\{\{x_1^1,...,x_1^k\}...\{x_i^1,...,x_i^k\}\}$. Meanwhile, the original logical label space corresponding to the whole is $L$, the label space of $i$-th bag can be denoted as $l_i$. $t$-th label of $i$-th bag can be denoted as $l_i^t$, the logical label space of the $i$-th bag can be denoted as $l_i=\{l_i^1,l_i^2,...,l_i^t\}$, therefore, the global logical label space of all bags also can be expressed as $L=\{\{l_1^1,...,l_1^t\}...\{l_i^1,...,l_i^t\}\}$. The global label distribution space of all bags is $D$, where label distribution space of the $i$-th bag can be denoted as $d_i$, therefore, the global label distribution space of all bags can also be represented as $D=\{d_1^1,...,d_1^t\}...\{d_i^1,...,d_i^t\}\}$.

The GLEMIML model is divided into three parts. The first part is a graph-based label enhancement. The second part is a MIML classifier with a simple two-layer fully connected network. Moreover, the third part is an interaction loss optimization for MIML enhancers and MIML Classifiers. To make the results of label enhancement from being effectively utilized by classifiers, GLEMIML uses the interactive loss optimization framework to guide the training of label enhancement using a MIML classifier. The flow chart of the model is shown in Figure 2.

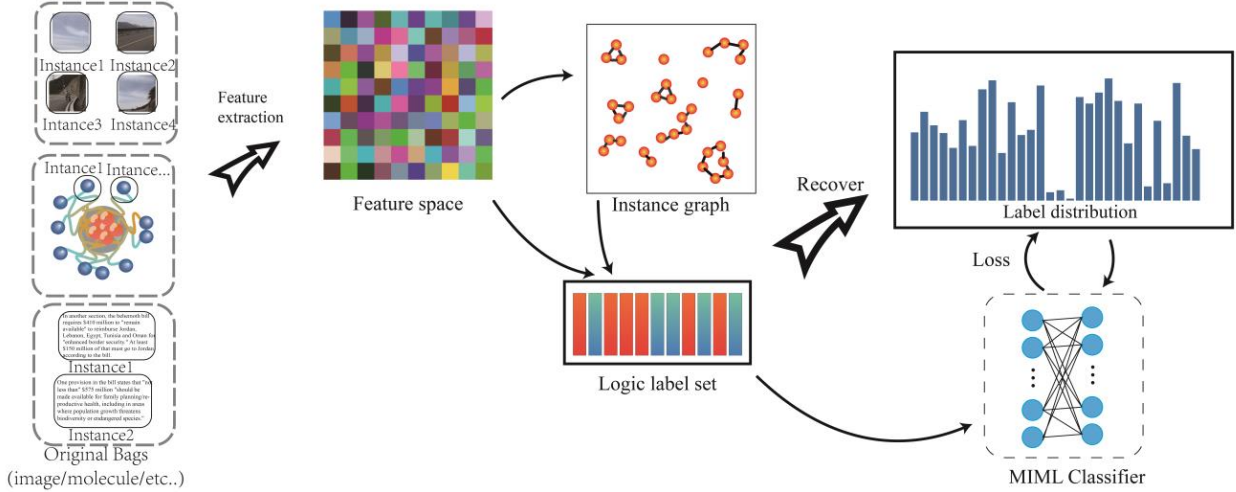

Figure 2: Flowchart of the proposed GLEMIML method. By mining the correlation between instances, topological information of feature space, and correlation between labels, the label space is recovered. Then a MIML classifier is used to classify multiple labels using the recovered label distribution and logical labels, and mixed loss is used to assist the learning of label distribution.

### 3.2 Label Enhancement

In MIML, with difficulty and cost in quantifying the label distribution, a large number of objects cannot be separated from instances in practical applications, such as protein molecules, complex images, etc., so one bag is often assigned with logic labels. The label distribution information can be recovered by mining relevant information in the feature space. In GLEMIML, label distribution information is recovered from logical labels by mining the correlation between labels[Tsoumakas et al., 2009], the correlation between instances and the topology information of feature space.

**Instance Relevance Mining**

In MIML tasks, many scenarios require multiple instances acting on same labels [Zhang and Zhang, 2006]. Compared with single instance multiple labels task, the sample of MIML may consist of multiple instances, Multiple instances often have correlations. However, traditional LE algorithms do not have the concept of multiple instances, and the correlation information between instances cannot be mined effectively.

At the same time, multiple instances often bring a large and redundant feature space, and directly mining the correlation of instances in the original feature space will bring a huge amount of computation and noise interference.

In order to effectively mine the correlation between instances, we first use the bag $x_i$ to project the data into a low-dimensional space in units of instances, which can be expressed as

$$\sigma(x_i)=\{\sigma(x_i^1),\sigma(x_i^2),...,\sigma(x_i^k)\} \quad (1)$$

where $\sigma(x_i)$ is a nonlinear projection that sequentially projects the instances in the bag into a low-dimensional feature space. In this way, according to the smoothing assumption [Zhu, 2005], two instances that are close to each other in the low-dimensional feature space may have correlation to the same labels. Therefore, define $\sigma(x_i^k)$ to be a nonlinear mapping of the $k$-th instance of the $i$-th bag. If $\sigma(x_i^k)$ is the K-nearest neighbor of $\sigma(x_i^m)$ and $\sigma(x_i^m)$ also is the K-nearest neighbor (KNN) of $\sigma(x_i^k)$, then $\sigma(x_i^m)$ is connected to $\sigma(x_i^k)$. Therefore, any instances in the bag can be expressed as

$$a_{km}=\begin{cases} 0, \sigma(x_i^k) \text{ and } \sigma(x_i^m) \text{ not KNN} \\ exp\left(\frac{|\sigma(x_i^k)-\sigma(x_i^m)|^2}{2\partial}\right), \sigma(x_i^k) \text{ and } \sigma(x_i^m) \text{ KNN} \end{cases} \quad (2)$$

where $\partial$ is the width parameter for similarity calculation.

A connected graph is established according to KNN, where the interaction function $W(x_i)$ between instances can be written as

$$W(x_i)=\sum_{k,m} a_{km}\sigma(x_i^k)G\sigma(x_i^m)^T \quad (3)$$

where $G$ is the Laplacian matrix [Merris, 1994] of the graph established by $x_i$ through the KNN. The above formula can effectively supplement the interaction between examples.

**Label Distribution Space Recovery**

Label enhancement is to recover label distribution by transferring the implicit information mined in feature space to label space. In the MIML task, it is the key to recovering label distribution space by mining the topological information of feature space, the correlation between instances, and the correlation between labels. Therefore, the formula for recovering label space can be expressed as

$$D=\omega_1(X)+\omega_2(W(X))+\omega_3(Y) \quad (4)$$

Where, $\omega_1(X)$ is the nonlinear mapping of feature topological information to mine the correlation of internal features of the instance; $\omega_2(W(X))$ is to mine the graph structure between instances and map it nonlinear to mine the correlation between instances; $\omega_3(Y)$ is the non-linear mapping of logical labels used to mine the correlation of label space. Finally, it is fused in the high dimensional space and migrated to the label space to recover the label distribution space.

| Dataset | instances | labels | Instances per bag (Mean± std.) | Labels per bag (Mean± std.) |
|---|---|---|---|---|
| Text Data For MIML(Text) | 2000 | 7 | 1.15 ± 0.37 | 3.56 ± 2.71 |
| Image Data for MIML(Image) | 2000 | 5 | 15.00±0.00 | 1.24±0.17 |
| Geobacter Sulfurreducens(GS) | 379 | 320 | 3.20±1.21 | 3.14±3.33 |
| Azotobacter Yinelandii(AV) | 407 | 340 | 3.07±1.16 | 4.00±6.97 |
| Haloarcula Marismortui(HM) | 304 | 234 | 3.13±1.09 | 3.25±3.02 |
| Pyrococcus Furiosus(PF) | 425 | 321 | 3.10±1.09 | 4.48±6.33 |
| Saccharomyces Cerevisiae(SC) | 3509 | 1566 | 1.86±1.36 | 5.89±11.52 |
| Caenorhabditis Elegans(CE) | 2512 | 940 | 3.39±4.20 | 6.07±11.25 |
| Drosophila Melanogaster(DM) | 2605 | 1035 | 3.51±3.49 | 6.02±10.24 |

Table 1: Statistics of the nine datasets.

**Label Correlation In Label Distribution**

Label correlation is also an implicit important information. In logical labels, the simplification is for {0,1} labels, and the correlation between labels is also simplified. For example, two labels may show different significance in the label distribution space, but in logical label space, they might all show up as 1, and their correlation is strengthened because of the simplification of logical labels. Thus, the correlation of relevant labels that have similar significance is diluted. Therefore, only mining label correlation in logical label space cannot effectively mine label correlation.

And when the labels are recovered from logical labels to label distribution, the inter-labels correlation information is also recovered. The closer the two labels are, the closer the descriptiveness corresponding to the labels should be. Therefore, the label distribution can be labeled as

$$D=T(D)+\omega_1(X)+\omega_2(W(X))+\omega_3(Y) \quad (5)$$

Among them $T(D)$, is to find the closest K related labels in the label distribution through KNN, and establish a connected graph between label.

### 3.3 Optimization Framework

It is important to avoid the separation of the classifier and label enhancer, because in this way, the label distribution recovered by the label enhancer may not be an effective guide to MIML classification. Not only that, the prediction information of the classifier guides the training of the label enhancer.

In order to avoid the separation of label enhancer and classifier in the training process, we use matching and interaction mechanism to realize the interaction between label enhancement and classifier.

**Label Enhancement Optimization Framework**

To recover a more accurate label distribution, the recovered label distribution can help the MIML classifier simultaneously. We construct a hybrid loss function optimizer. The label enhancement loss function of the label enhancer is represented by the prediction loss of the MIML classifier, bag correlation loss, and the loss caused by the threshold strategy.

Inspired by the asymmetric focal loss [Ridnik et al., 2021] [Lin et al., 2017], we use the same interaction label loss to construct the prediction loss of the MIML classifier to guide the optimization of label enhancement.

$$L_{CL}=-\frac{1}{k}\sum_{j=1}^{k}\begin{cases}(1-p_j)^{\gamma+} log(p_j^*), j \in \Omega_{pos} \\ (p_j)^{\gamma-} log(1-p_j^*), j \in \Omega_{neg}\end{cases} \quad (6)$$

where $p_j$ is the predicted output of the prediction model via sigmoid, and $p_j^*$ is the predicted output of the label distribution loss estimate via the sigmoid operation. $\Omega_{pos}$ is the set of related samples, and $\Omega_{neg}$ is the set of uncorrelated samples. $\gamma+$ and $\gamma-$ are two hyperparameters used to control the loss function.

For label enhancement, according to the smoothness assumption, we assume that two models with similar feature spaces can perform similar label spaces.

$$Z_{ij}=sim(x_i,x_j) \quad (7)$$

$sim(a,b)$ is a metric similarity function, i.e., cosine similarity.

With similar feature spaces and topologies in two bags, the label distribution spaces of them could maintain similar distributions.

$$A_{ij}=sim(d_i,d_j) \quad (8)$$

Where, $d_i$ and $d_j$ are the corresponding label distribution values of $x_i$ and $x_j$.

Therefore, the loss function for bag correlation can be expressed as

$$L_{Sim}=\left(\frac{\sum_i^m \sum_j^i (Z_{ij}-A_{ij})}{Z}\right)^2 \quad (9)$$

where $Z$ is the number of bags.

Because the label enhancement is data preprocessing rather than data noise detection and correction, we assume that the labeling significance of relevant labels in logical labels must be greater than that of irrelevant labels. To avoid when after label enhancement, the contiguous values of relevant labels are lower than those of irrelevant labels. Therefore, in the training process, we must set a threshold to make relevant labels more descriptive than irrelevant labels.

The loss function based on the threshold strategy can be expressed as

$$L_{threshold}=\frac{1}{M}\sum_{i=1}^{M} max(max(d_i^{neg}) - min(d_i^{pos}),0) \quad (10)$$

among them, $d_i^{pos}$ refers to the set of label distribution values corresponding to all relevant labels in the $i$-th bag, and $d_i^{neg}$ refers to the set of label distribution values corresponding to all irrelevant labels in the $i$-th bag.

| Dataset | Metrics | LE | | MIML | | | | | |
|---|---|---|---|---|---|---|---|---|---|
| | | GLEMIML | MI-FLEM | MIMLNN | MIMLSVM | EnMIMLNN | WEL | KASIR | |
| Image | HL↓ | **0.1650(1)** | 0.2480(5) | 0.2252(4) | 0.3408(7) | 0.1736(2) | 0.3275(6) | 0.1867(3) | |
| | RL↓ | **0.1590(1)** | 0.2635(5) | 0.2513(4) | 0.4723(7) | 0.1900(3) | 0.4698(6) | 0.1780(2) | |
| | mAP↑ | 0.7091(3) | 0.4938(7) | 0.7085(4) | 0.5084(6) | 0.7652(2) | 0.6132(5) | **0.8012(1)** | |
| | Ma-F1↑ | **0.6747(1)** | 0.4693(7) | 0.6111(4) | 0.6441(3) | 0.5975(5) | 0.5434(6) | 0.6617(2) | |
| Text | HL↓ | **0.0235(1)** | 0.0332(2) | 0.0384(3) | 0.1753(7) | 0.0457(5) | 0.1143(6) | 0.0420(4) | |
| | RL↓ | **0.0097(1)** | 0.0156(2) | 0.0261(3) | 0.2273(7) | 0.0372(5) | 0.2130(6) | 0.0288(4) | |
| | mAP↑ | **0.9467(1)** | 0.8980(5) | 0.9203(4) | 0.6668(7) | 0.9390(3) | 0.8568(6) | 0.9418(2) | |
| | Ma-F1↑ | **0.9380(1)** | 0.9119(3) | 0.8804(4) | 0.9201(2) | 0.8484(6) | 0.8371(7) | 0.8674(5) | |
| GS | HL↓ | **0.0089(1)** | 0.0098(3) | 0.0118(6) | 0.0111(5) | 0.0097(2) | 0.0126(7) | 0.0099(4) | |
| | RL↓ | **0.2847(1)** | 0.3309(4) | 0.3415(5) | 0.2946(2) | 0.3175(3) | 0.7167(6) | 0.8724(7) | |
| | mAP↑ | 0.2353(3) | 0.0911(6) | 0.2443(2) | 0.2251(4) | **0.3397(1)** | 0.2094(5) | 0.0221(7) | |
| | Ma-F1↑ | **0.0750(1)** | 0.0272(4) | 0.0040(7) | 0.0050(6) | 0.0669(2) | 0.0428(3) | 0.0147(5) | |
| AV | HL↓ | **0.0088(1)** | 0.0114(3) | 0.0126(4) | 0.0147(5) | 0.0107(2) | 0.0148(6) | 0.0150(7) | |
| | RL↓ | 0.3271(2) | 0.3947(4) | 0.4095(5) | **0.2408(1)** | 0.3329(3) | 0.7655(6) | 0.9129(7) | |
| | mAP↑ | **0.2628(1)** | 0.1069(6) | 0.1805(5) | 0.2612(3) | 0.2623(2) | 0.1909(4) | 0.0330(7) | |
| | Ma-F1↑ | **0.0789(1)** | 0.0645(2) | 0.0053(7) | 0.0064(6) | 0.0475(4) | 0.0445(5) | 0.0157(5) | |
| HM | HL↓ | **0.0109(1)** | 0.0128(3) | 0.0153(6) | 0.0142(4) | 0.0119(2) | 0.0162(7) | 0.0147(5) | |
| | RL↓ | **0.2209(1)** | 0.3524(5) | 0.2834(4) | 0.2352(2) | 0.2644(3) | 0.6006(6) | 0.8529(7) | |
| | mAP↑ | 0.3236(2) | 0.2434(6) | 0.2577(5) | 0.2890(3) | **0.4201(1)** | 0.2679(4) | 0.0541(7) | |
| | Ma-F1↑ | **0.1585(1)** | 0.1126(3) | 0.0072(7) | 0.0239(6) | 0.0732(4) | 0.1149(2) | 0.0430(5) | |
| PF | HL↓ | **0.0084(1)** | 0.0129(2) | 0.0136(3) | 0.0154(4) | 0.0160(6) | 0.0187(7) | 0.0154(4) | |
| | RL↓ | **0.2407(1)** | 0.3073(4) | 0.3493(5) | 0.2886(3) | 0.2687(2) | 0.6006(6) | 0.8420(7) | |
| | mAP↑ | 0.2690(2) | 0.1159(6) | 0.2364(4) | 0.2384(3) | **0.3714(1)** | 0.2132(5) | 0.0657(7) | |
| | Ma-F1↑ | **0.1138(1)** | 0.0430(4) | 0.0062(7) | 0.0080(6) | 0.0734(2) | 0.0268(5) | 0.0468(3) | |
| SC | HL↓ | **0.0035(1)** | 0.0039(3) | 0.0041(5) | 0.0041(5) | 0.0036(2) | N/A(7) | 0.0039(3) | |
| | RL↓ | **0.2150(1)** | 0.2230(2) | 0.2363(4) | 0.2192(2) | 0.3209(5) | N/A(7) | 0.8437(6) | |
| | mAP↑ | 0.0428(4) | 0.0373(5) | 0.1528(3) | 0.1931(2) | **0.1950(1)** | N/A(7) | 0.0279(6) | |
| | Ma-F1↑ | 0.0123(2) | 0.0015(6) | 0.0061(4) | 0.0021(5) | 0.0114(3) | N/A(7) | **0.0243(1)** | |
| CE | HL↓ | **0.0054(1)** | 0.0057(3) | 0.0071(6) | 0.0060(5) | 0.0055(2) | 0.0094(7) | 0.0058(4) | |
| | RL↓ | **0.1335(1)** | 0.2088(5) | 0.2051(4) | 0.1622(2) | 0.1641(3) | 0.5733(6) | 0.7344(7) | |
| | mAP↑ | 0.2598(5) | 0.2662(4) | 0.1983(6) | 0.4309(2) | **0.5216(1)** | 0.3149(3) | 0.1277(7) | |
| | Ma-F1↑ | **0.1576(1)** | 0.1324(2) | 0.0277(7) | 0.0301(6) | 0.1116(4) | 0.0948(5) | 0.1256(3) | |
| DM | HL↓ | **0.0047(1)** | 0.0085(7) | 0.0062(4) | 0.0055(3) | 0.0063(5) | 0.0081(6) | 0.0053(2) | |
| | RL↓ | **0.1636(1)** | 0.1831(4) | 0.2058(5) | 0.1730(2) | 0.1820(3) | 0.5843(6) | 0.7072(7) | |
| | mAP↑ | 0.4625(2) | 0.2225(5) | 0.1797(6) | 0.3769(3) | **0.4969(1)** | 0.2857(4) | 0.1601(7) | |
| | Ma-F1↑ | **0.2463(1)** | 0.1508(3) | 0.0252(6) | 0.0250(7) | 0.1122(5) | 0.1234(4) | 0.1587(2) | |
| Avg. Rank | | 1.44 | 4.19 | 4.78 | 4.25 | 2.92 | 5.56 | 4.78 | |

Table. 2 GMIML was compared with other MIML models and the LE model MI-FLEM. Due to the large sample size of SC data set, MIMLWEL cannot operate effectively under 32G memory. We believe that the lack of this experimental result will not affect the conclusion of the experiment, so we give up the collection.

Therefore, our final optimization framework can be expressed as

$$L_{CLE}=\beta_1 L_{CL}+\beta_2 L_{Sim}+\beta_3 L_{threshold}, \sum_{i=1}\beta_i =1 \quad (11)$$

among them, $\beta_1$, $\beta_2$ and $\beta_3$ are a set of hyperparameters, which are used to optimize the ratio of the three strategies of classifier, threshold and similarity between bags.

**Classifier Optimization Framework**

While using the classification output to guide the training of label enhancement, it is also possible to apply the label distribution to guide the learning of the classifier. Compared with MIMIL's algorithm, leveraging label distribution can learn relevant features more effectively.

For the loss function that the MIML classifier uses label distribution to guide classification, we define it as follows

$$L_{DC}=\frac{1}{n}\sum_{i=1}^{n}(\sum_{j=1}^{k} d_i^j log d_i^j \left( e^{\sum_{u=1}^{k} s_i^u - s_i^j} \right)) \quad (12)$$

among them, $n$ is the number of bags, and $k$ is the number of labels. $s_i^j$ is the logical prediction output of the classifier for the $i$-th bag. The uniform label distribution loss $L_{DC}$ is a generalized form of cross-entropy with the upper bound of the cross-entropy loss, which can effectively help the model to achieve better convergence [Zhao et al., 2022].

Additionally, the training mechanism of matching interaction makes the output of enhanced labeling cannot effectively guide the classifier learning at the beginning, so the loss of classification output and logical labeling is introduced to guide the learning of the classifier. Here we choose a binary cross-entropy loss function.

$$L_{LC}=\frac{1}{c}\sum_{j=1}^{c} \begin{cases} log(p_j), j \in \Omega_{pos} \\ log(1-p_j), j \in \Omega_{neg} \end{cases} \quad (13)$$

The final loss function of our proposed method can be obtained as follows

$$L_C=\rho L_{LC}+(1-\rho)L_{DC} \qquad (14)$$

where $\rho$ is a hyperparameter that balances the logistic loss and label distribution loss.

## 4 Experiments

### 4.1 Datasets & Features

In this section, nine real-world MIML datasets are used for experimental comparisons, including one image dataset for MIML [Boutell et al., 2004], one text dataset for MIML [Zhang and Zhou, 2008], and other seven protein MIML datasets [Apweiler et al., 2004]. We divide the data set into the training set, test set and verification set according to 7:2:1, and the statistics of the final data set are given in Table 1.

### 4.2 Comparing Algorithms And Evaluation

Five state-of-the-art MIML algorithms include MIMLNN [Chen et al., 2013], MIMLSVM [Zhou, 2012], MIMLWEL [Yang et al., 2013], KASIR[Li et al., 2012], EnMIMLNN [Wu et al., 2014] are selected for comparisons. Additionlly, since there are no related MIML algorithms considering label enhancement, to prove the effectiveness of our proposed GLEMIML, one representative label enhancement algorithm namely FLEM [Zhao et al., 2022] is also used for comparison, which is trained by the matching interaction mechanisms will expanding all examples into one-dimensional vectors in order as input, thus making FLEM can achieve MIML using label enhancement with namely MI-FLEM. All the parameters are set by default parameters of the original paper.

Due to the limitations, the four most commonly used evaluation metrics in label enhancement are selected for experimental analysis, including Hamming Loss (HL), Ranking Loss (RL), macro-Average Precision (mAP), and Ma-Fl [Yang et al., 2013] [Ghamrawi and McCallum, 2005] [Rogati and Yang, 2002], The smaller value the first two metrics, the better performances can be obtained. And the larger value of the latter two metrics, the better performance.

### 4.3 Comparing Experimental Results

As shown in Table 2, the best performance is marked in bold. Meanwhile, we also show the results of the average rank of each comparing method. By analyzing the experimental results, we can get the following conclusions. 1) Among the 36 cases (9 datasets × 4 evaluation metrics), our model has reached the first rank in the overall ranking, and it outperforms the other comparing algorithms in most cases 2) Compared with the traditional MIML algorithm, the two MIML algorithms by considering label enhancement have reached the first or third place in the comprehensive ranking, which shows that effectiveness and significance of leveraging label significances in the MIML task. 3) Nevertheless, compared with the FLEM, our proposed model shows better performance. With a more complex feature space in MIML, FLEM cannot fully and effectively mine the hidden information from the feature space, while GLEMIML fully exploits the latent information to supply a more effective and accurate model of label enhancement, thus achieving more satisfactory performances.

| Scenario | Metrics | Image | SC |
|---|---|---|---|
| GLEMIML | HL↓ | 0.1650 | 0.0035 |
| | RL↓ | 0.1590 | 0.2150 |
| | mAP↑ | 0.7091 | 0.0428 |
| | Ma-F1↑ | 0.6747 | 0.0123 |
| GLEMIML-A | HL↓ | 0.2650 | 0.0037 |
| | RL↓ | 0.3072 | 0.2503 |
| | mAP↑ | 0.4451 | 0.0147 |
| | Ma-F1↑ | 0.4156 | 0.0004 |
| GLEMIML-B | HL↓ | 0.1700 | 0.0032 |
| | RL↓ | 0.1735 | 0.2261 |
| | mAP↑ | 0.6930 | 0.0492 |
| | Ma-F1↑ | 0.6388 | 0.0030 |
| GLEMIML-C | HL↓ | 0.1845 | 0.0037 |
| | RL↓ | 0.1877 | 0.2429 |
| | mAP↑ | 0.6338 | 0.0380 |
| | Ma-F1↑ | 0.5125 | 0.0089 |

Table 3: Ablation experiment

### 4.3 Ablation Study

Ablation experiments of GLEMIML on two representative datasets are further conducted to demonstrate the effectiveness of each step in our model. The experimental results are shown in Table 3.

GLEMIML-A is a model that chooses a single-layer fully connected neural network as a classifier; GLEMIML-B is a model that chooses a three-layer fully connected neural network as a classifier; GLEMIML-C cancels the interaction between instances in the model. Through the comparison of GLEMIML with GLEMIML-A and GLEMIML-B, although the selection of a single-layer fully connected classifier can also guide the learning of MIML, simple linear classification cannot achieve more complex classification tasks. When the number of layers of the classifier exceeds two layers, although multiple layers may bring relatively better extraction results, there are no longer better benefits while with a cost of converging not easily. Compared with GLEMIML-C without considering the instance's correlation, GLEMIML fully mined the correlation between instances, thus improving the effectiveness of label enhancement for MIML tasks.

## 5 Conclusions

This paper proposes a novel joint MIML framework based on graph label enhancement, namely GLEMIML, to unravel the problems in which ignoring labeling significance easily results in ineffective learning problems in MIML. GLEMIML uses the correlation graph structure to mine the correlation between instances to realize the label distribution by migrating the feature topology information to the label space. Subsequently, by utilizing a MIML classifier to guide the training of LE through the matching and interaction mechanism, GLEMIML can achieve more effective and accurate performances by leveraging label distribution. The experimental results on multiple datasets prove the superiority of GLEMIML on the MIML task. GLEMIML (AvgRank: 1.44) performed much better in the data set than MI-FLEM(Avg. Rank: 4.19) compared to improved SML markup enhancement. Compared with the multi-example multi-label algorithm, GLEMIML(AvgRank: 1.44) is also much higher than the optimal model EnMIMLNN(Avg. Rank: 2.92).